\documentclass[conference]{IEEEtran}
\IEEEoverridecommandlockouts
\usepackage{cite}
\usepackage{amsmath,amssymb,amsfonts}
\usepackage{algorithmic}
\usepackage{graphicx}
\usepackage{textcomp}
\usepackage{xcolor}
\usepackage{adjustbox}
\usepackage{subcaption}
\usepackage{multirow, multicol}
\usepackage{booktabs}

\usepackage{pifont}
\newcommand{\cmark}{\ding{51}}%
\newcommand{\xmark}{\ding{55}}%

\usepackage[nolist]{acronym}
\def\BibTeX{{\rm B\kern-.05em{\sc i\kern-.025em b}\kern-.08em
    T\kern-.1667em\lower.7ex\hbox{E}\kern-.125emX}}
\begin{document}

\title{Efficient Defense Against Model Stealing Attacks on Convolutional Neural Networks
\thanks{
This research was funded by Synopsys Inc. and the Natural Sciences and Engineering Research Council of Canada (NSERC).
}
}

\author{


\IEEEauthorblockN{
Kacem Khaled\IEEEauthorrefmark{1}, 
Mouna Dhaouadi\IEEEauthorrefmark{2}, 
Felipe Gohring de Magalhães\IEEEauthorrefmark{1} and 
Gabriela Nicolescu\IEEEauthorrefmark{1}
}
\IEEEauthorblockA{
\IEEEauthorrefmark{1}Department of Computer Engineering and Software Engineering,
Polytechnique Montreal, Canada\\
\IEEEauthorrefmark{2}Department of Computer Science and Operations Research,
University of Montreal, Canada\\
Email: kacem.khaled@polymtl.ca
}


}

\IEEEoverridecommandlockouts
\IEEEpubid{\makebox[\columnwidth]{To appear in Proceedings of ICMLA, Florida, USA
\copyright2023
IEEE\hfill} \hspace{\columnsep}\makebox[\columnwidth]{}}

\maketitle

\begin{abstract}

Model stealing attacks have become a serious concern for deep learning models, where an attacker can steal a trained model by querying its black-box API. This can lead to intellectual property theft and other security and privacy risks. 
The current state-of-the-art defenses against model stealing attacks suggest adding perturbations to the prediction probabilities. However, they suffer from heavy computations and make impracticable assumptions about the adversary. They often require the training of auxiliary models. This can be time-consuming and resource-intensive which hinders the deployment of these defenses in real-world applications.
In this paper, we propose a simple yet effective and efficient defense alternative. We introduce a heuristic approach to perturb the output probabilities. 
The proposed defense can be easily integrated into models without additional training. 
We show that our defense is effective in defending against three state-of-the-art stealing attacks.
We evaluate our approach on large and quantized (i.e., compressed) Convolutional Neural Networks (CNNs) trained on several vision datasets. 
Our technique outperforms the state-of-the-art defenses with a $\times37$ faster inference latency without requiring any additional model and with a low impact on the model's performance. We validate that our defense is also effective for quantized CNNs targeting edge devices. 

\end{abstract}


\newcommand\kacem[1]{\textcolor{blue}{#1}}

\newcommand\felipe[1]{\textcolor{red}{#1}}

\newcommand\mouna[1]{\textcolor{green}{#1}}

\newcommand\rewrite[1]{\textcolor{gray}{#1}}

\begin{acronym}
  \acro{ML}{Machine Learning}
  \acro{DL}{Deep Learning}
  \acro{MLaaS}{Machine Learning as a Service}
  \acro{ANN}{Artificial Neural Networks}
  \acro{API}{Application Programming Interface}
  \acro{CNN}{Convolutional Neural Network}
  \acro{CNNs}{Convolutional Neural Networks}
  \acro{DNN}{Deep Neural Network}
  \acro{DNNs}{Deep Neural Networks}
  \acro{NN}{Neural Network}
  \acro{ART}{Adversarial Robustness Toolbox }
  \acro{DoS}{Denial-of-Service}
  \acro{FGSM}{Fast Gradient Sign Method}
  \acro{PGD}{Projected Gradient Descent}
  \acro{MLP}{Multilayer Perceptron}
  \acro{IP}{Intellectual Property}
  \acro{NLP}{Natural Language Processing} 
  \acro{PTQ}{Post Training Quantization} 
  \acro{QAT}{Quantization Aware Training} 
  \acro{RS}{Reverse Sigmoid} 
  \acro{AM}{Adaptive Misinformation} 
  \acro{DCP}{Deception} 
  \acro{CDCP}{Combined Deception}
  \acro{Soft-RS}{Softplus Reverse Sigmoid}
  \acro{OOD}{Out-of-Distribution}
  \acro{SGD}{Stochastic Gradient Descent}
\end{acronym}

\section{Introduction}

\label{sec:sota}

\ac{DL} achieved human-level performance in several computer vision tasks~\cite{krizhevsky2017imagenet}. Developing such high-performance models is a costly process for  companies~\cite{strubell2019energy}. 
They often provide their models as a service through an \ac{API}. This enables users to have predictions for their queries. Besides, state-of-the-art \ac{DNNs} are more and more computationally expensive and require intensive memory, which in some cases hinder their deployment in embedded devices. Therefore, during the past years, researchers proposed to \emph{quantize} \ac{DNNs} without significantly affecting their performance~\cite{survey_q}.
This enables a \ac{DL} model to run efficiently on edge devices to enable a service for a user (e.g.~smartphones) or other applications in the system (e.g.~autonomous vehicle).

Sharing models as a black box to users or between companies creates a dilemma for the model owner since there might be a financial risk of \ac{IP} theft.
Some research work focuses on obfuscating the model architecture to hinder attackers from retrieving the original model architecture. 
Nevertheless, the model obfuscation is useless when the attacker can retrieve a surrogate model using only the prediction \ac{API}~\cite{tramer2016stealing}. 
An attacker can act as a normal user and send input queries to the \ac{DL} model. Leveraging the output predictions, a malicious user can train a high-accuracy and high-fidelity copy of the original model.
In addition, model stealing attacks can be used to facilitate other attacks such as adversarial attacks where the attacker creates malicious examples to evade the model classification~\cite{papernot2017practical}.


Previous work proposed different defenses against API-based model stealing attacks.
Some work proposes to withhold information about probabilities, but this reduces the model's transparency and utility~\cite{tramer2016stealing}.
The prediction confidence is important information, especially for critical systems, e.g. in an autonomous vehicle the system might take an action only when a confidence threshold is reached.
Other work proposes to add perturbations to the prediction probabilities in a way that reduces the attacker performance~\cite{Orekondy2019poisoning,mazeika2022steer,lee2018defending}.
However, these techniques often inject high perturbation which hurts the model transparency or even reduces its performance. 
Additionally, state-of-the-art defense techniques are computationally expensive. 
Despite the focus of previous work on maintaining the defender model's performance by adding a minimal perturbation, none of them measure their defenses' impact on the model inference latency~\cite{Orekondy2019poisoning,mazeika2022steer,lee2018defending}.
In critical systems, inference latency is very important which makes state-of-the-art defenses infeasible. 
When developing \ac{DL} models targeting edge devices, it is crucial to have 
efficient algorithms
that neither overload the system nor reduce its utility~\cite{survey_q}.

To overcome these issues, we propose a novel efficient perturbation-based defense alternative called \emph{\ac{DCP}}.
Our \emph{heuristic}  approach aims to deceive the attacker and decrease the accuracy of his stolen model by adding an adaptive noise that deviates the probability scores without increasing the inference time. In order to keep the model's transparency, we aim to decrease the attack performance with a low amount of perturbation.
To the best of our knowledge, our approach is the first to propose a defense that takes into account the \textit{attack performance}, the \textit{defender's accuracy}, the \textit{perturbation level}, the \textit{inference time}, the \textit{energy consumption} and the \textit{number of required models} to perform the defense (Table~\ref{tab:compare}).
We test our approach on several vision datasets: CIFAR-10, CIFAR-100~\cite{krizhevsky2009cifar10},  SVHN~\cite{netzer2011svhn}, GTSRB~\cite{Stallkamp2011gtsrb} and CUB200~\cite{WahCUB_200_2011}.
Our evaluation proves that our technique is effective in defending against three state-of-the-art model stealing attacks \emph{KnockoffNets}~\cite{orekondy2019knockoff}, \emph{MAZE}~\cite{kariyappa2021maze} and \emph{DFME}~\cite{truong2021data}.
We find better or comparable results with state-of-the-art defenses in the matter of the perturbation budget and the attack performance. 
Furthermore,
our technique outperforms the state-of-the-art 
defenses
\emph{GRAD$^2$}~\cite{mazeika2022steer} and \emph{MAD}~\cite{Orekondy2019poisoning}
with a faster inference latency by consuming less energy and without relying on other surrogate models.
We validate our approach on \emph{large} (i.e., original architecture) and \emph{quantized} \ac{CNNs}. 
We show that the latter is as vulnerable to model stealing attacks as their corresponding original models. 
Our technique succeeds in defending against these attacks in the quantization setting which ensures that our defense extends to real-world applications in edge devices. 
Our code is available at: 
https://github.com/KacemKhaled/defending-extraction




The remainder of the paper is organized as follows: 
Section~\ref{sec:sota} reviews the state-of-the-art that relates to model stealing attacks and defenses, and CNNs quantization; 
Section~\ref{sec:methodology} details our methodology; 
we include our experiments and obtained results in Section~\ref{sec:experiments}; 
and Section~\ref{sec:conclusion} concludes the paper.

\begin{table}[]
\centering
\caption{Our approach considers multiple aspects while defending against model stealing attacks}
\label{tab:compare}
\begin{tabular}{cccc}
\toprule
Approach & {Fast inference} & {No auxiliary model} & {Low perturbation} \\ 
\midrule
RS \cite{lee2018defending}      & \cmark & \cmark  & \xmark  \\ 
AM \cite{kariyappa2020defending}& \cmark & \xmark   & \cmark  \\ 
MAD \cite{Orekondy2019poisoning}& \xmark & \xmark & \cmark  \\ 
GRAD$^2$ \cite{mazeika2022steer}   & \xmark & \xmark & \cmark  \\ 
\textbf{Ours}                   & \cmark & \cmark & \cmark  \\ 
\bottomrule
\end{tabular}
\end{table}


\section{Related Work}
\label{sec:sota}


\subsection{Model stealing attacks}

Model stealing attacks also referred to as model extraction attacks, can be categorized into two groups: attacks that exploit hardware access and attacks that leverage \ac{API} query access. We refer to the latter category with API-based attacks. 
In API-based  attacks, the adversary observes the prediction outputs of his queries, then uses them  to steal the functionality of a model~\cite{tramer2016stealing}. 
Prior works \cite{papernot2017practical,Juuti2019} propose extraction attacks using a jacobian-based data augmentation approach to produce synthetic data. However, their success is limited to small datasets.
Reference \cite{orekondy2019knockoff} proposes \emph{KnockoffNets} attack where the adversary aims to obtain a functionally equivalent clone of the victim model using random and publicly available data. The attack is easy and simple to execute, yet very effective in outperforming other state-of-the-art stealing attacks.
Recently, other works in extraction attacks namely \emph{MAZE}~\cite{kariyappa2021maze} and \emph{DFME}~\cite{truong2021data} focus on modern techniques to generate the adversary's dataset instead of relying on public datasets. They leverage generative models to generate data with an objective that enables a successful extraction. However, they require millions of queries instead of thousands compared to \textit{KnockoffNets}~\cite{orekondy2019knockoff}.
Other stealing attacks in the literature assume that a target model can only output hard labels instead of probabilities~\cite{Zhou_2020_CVPR_dast,Sanyal_2022_CVPR_hard}.

\subsection{Model stealing defenses}

To enable  model owners to claim their models if stolen, previous work proposes watermarking the model \cite{jia2020entangled,szyller2019dawn}. 
Watermarking can be done by changing the output probabilities for a small subset of queries. However, this technique does not prevent model stealing.
Other defense approaches protect against attacks by perturbing the posterior prediction. 
These perturbation-based defenses can be categorized into two categories: \emph{proactive} and \emph{reactive} techniques.  

Reactive techniques are triggered with the detection of an ongoing extraction attack. 
They continuously monitor the interactions with the user and the defended model to be able to assess the knowledge stolen by the attacker~\cite{Kesarwani2018} or to detect an ongoing attack through detecting a deviation from the normal distribution~\cite{Juuti2019} or a large number of \ac{OOD} queries~\cite{kariyappa2020defending}. 
To detect \ac{OOD} queries, \cite{kariyappa2020defending} proposes a technique called \emph{\ac{AM}} where they flag as malicious queries the ones with low prediction scores. 
Then they use another model for ``misinformation'' that generates noise vectors which will be combined with the prediction posteriors. However, this technique is sensitive to false positives which deteriorates the model's performance. Reactive defenses  help reduce the efficiency of attacks, but they cause inference delays and increase computational costs.

Proactive defenses are always perturbing the posterior predictions, regardless an attack is detected or not. Some techniques propose to truncate the probability scores or to add random noise~\cite{orekondy2019knockoff}. But, these techniques are not efficient against state-of-the-art attacks.
The \emph{\ac{RS}} defense \cite{lee2018defending} leverages the last layer in a neural network, i.e., the one that outputs the prediction probabilities of each class, and changes its activation function to a reverse sigmoid. 
This technique uses a high perturbation level which hurts the model's transparency. To find the suitable perturbation per query, \emph{MAD}~\cite{Orekondy2019poisoning}  tries to solve an optimization problem where it simulates the attacker by another network and tries to add the perturbation that maximizes the error of the simulated surrogate model. It adds targeted noise to the posterior probabilities to poison the adversary's gradient signal. 
\emph{GRAD$^2$}\cite{mazeika2022steer} is a  similar approach to MAD that relies on redirecting the gradient signal. Besides, it proposes to train a surrogate model on the attacker's queries to simulate the attacker's knowledge. GRAD$^2$ always assumes that the attacker will send a large batch of queries that it can use for training the simulated attacker. This hinders its practicality since it assumes that the system, where the model is implemented to perform predictions, has sufficient hardware to perform the training.

In general, existing perturbation-based defenses are computationally expensive or they use high perturbations that hurt the model's transparency. However, in real-world systems that make decisions based on predictions (e.g., a computer vision system in an autonomous vehicle), the model has to provide a level of confidence in its predictions. The techniques that provide high perturbations make the model useless for these applications.
In addition, optimization-based techniques require a lot of gradient calculations, which results in enormous inference delays and an increase in computational costs.
Our proposed defense overcomes these issues through a heuristic approach that aims to add perturbations without requiring more hardware resources, as shown in Table~\ref{tab:compare}.   We offer better trade-offs in satisfying the perturbation budget vs. the defender's error constraint.

\subsection{\ac{CNNs} Quantization}
Quantization is a technique used to reduce the memory footprint and computation cost of \ac{CNNs} by representing their weights and activations using lower precision data types. 
In other words, quantizing the weights involves reducing the number of bits used to represent the parameters, typically from the standard 32-bit floating-point representation to 8-bit or even lower precision~\cite{han2015deep,vanhoucke2011improving}.
In this paper, we consider two of the most used quantization techniques for \ac{CNNs}: \ac{PTQ} and \ac{QAT}~\cite{survey_q}. 
    \ac{PTQ} finds the quantization parameters without retraining the model. It computes the resulting distributions of the different activation functions after feeding some batches of data to the trained model, to determine the specific quantization of each activation at inference time. Then the model is quantized based on the calibration results. This technique is fast and simple, however, the quantized model may have lower accuracy than the original one~\cite{survey_q}. 
   During \ac{QAT}, float parameter values are rounded to mimic int8 values, so the weight adjustments are made while being \textit{aware} that the model will be quantized. The calibration is performed while re-training the model. This technique often results in quantized models with higher accuracy than with \ac{PTQ}~\cite{survey_q}.

\section{Methodology}
\label{sec:methodology}
In this section, we present the threat model, the attack strategy, and our proposed defense approach.
\subsection{Threat Model}
\label{sec:threat-model}


\begin{figure}[]
\centering
\includegraphics[width=0.8\linewidth]{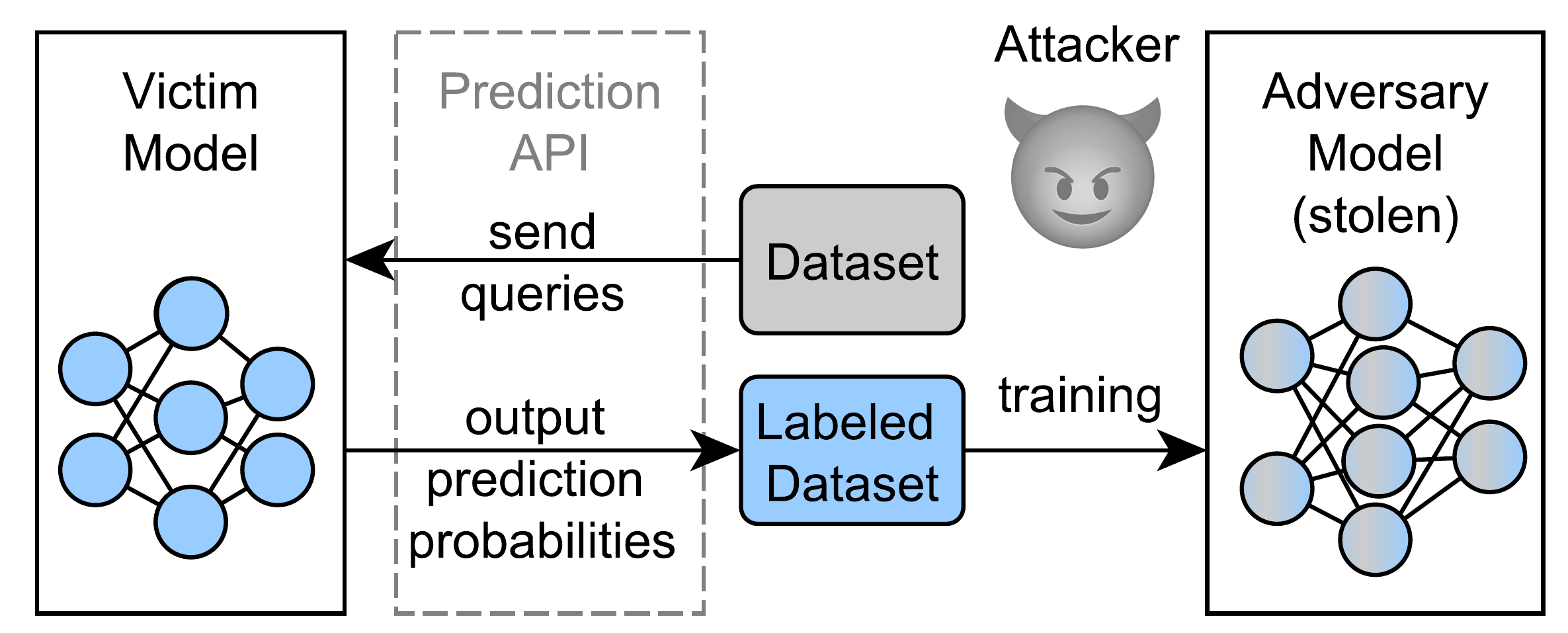}
\caption{In an API-based model stealing attack, the attacker sends queries to the victim model through its prediction API, then he uses the predictions received to create a labeled dataset. The attacker then trains a new model, called the \textit{Adversary} Model, on that dataset. This model represents the stolen model from the victim.}
\label{fig:attack}
\end{figure}

We assume that the model owner provides the \ac{DL} model as a black box.
The attacker has an \ac{API} access to the victim model without internal knowledge about the model parameters nor the training data.
The attacker is able to act as a normal user by sending queries to the model and receiving prediction probabilities.
Fig.~\ref{fig:attack} shows an API-based model stealing attack.
In these attacks, the attacker attempts to steal the functionality of the victim model by labeling a dataset with the prediction outputs obtained from the attacked model.
The goal of the attacker is to obtain a clone with high fidelity to the attacked model and high accuracy on the victim's prediction task.


\subsection{Attack strategy} 
Our aim is to comprehensively evaluate our perturbation-based defense approach against diverse strategies for model stealing.
We consider attacks that differ in the query distributions and strategies (random vs. adaptive querying).
Our defense aims to perturb the prediction probabilities, thus, we use the state-of-the-art model stealing attacks that leverage the output probabilities of target victim models: \emph{KnockoffNets}~\cite{orekondy2019knockoff}, \emph{MAZE}~\cite{kariyappa2021maze} and \emph{DFME}~\cite{truong2021data}.

Let $F_V(x;\theta)$ be the victim model provided by the prediction \ac{API}, with $x$ representing the input query and $\theta$ the model parameters.
The stealing attack approach has three steps: (i)~selecting a query set $\mathcal{Q}$, (ii)~constructing a transfer set, and (iii)~training a stolen model (the adversary's model) ${F_A}$.

MAZE~\cite{kariyappa2021maze} and DFME~\cite{truong2021data} attacks rely on generative models to create the query set $\mathcal{Q}$. They leverage the victim's outputs and estimate their gradients to adaptively construct suitable image queries. In these attacks steps (i) and (ii) happen simultaneously. 
KnockoffNets~\cite{orekondy2019knockoff} attack selects publicly available datasets to construct the query set $\mathcal{Q}$. 
The attacker selects an image distribution and randomly samples images from it.
We work with different datasets to simulate two scenarios: a \textit{knowledgeable} adversary and a \textit{limited-knowledge} adversary.
The former is an attacker with knowledge about the distribution of the training data. 
In this case, we select~$\mathcal{Q}$ that comes from the same distribution as the victim's dataset~$\mathcal{D}$, so there might be overlapping data between both datasets. We call this a distribution-aware attack.
For knowledge-limited adversary, we select a different dataset from the victim $\mathcal{Q} \neq \mathcal{D}$.
The adversary uses the obtained output predictions to label the data~$\mathcal{Q}$. Finally, using the obtained transfer set, the adversary  trains the stolen model ${F_A}$ using a reasonably complex architecture (e.g., ResNets~\cite{he2016deep}). 
The adversary's model ${F_A}$ can be trained to substitute ${F_V}$ by minimizing the cross-entropy loss, defined with $y_{i}=F_{V}(x;\theta)_{i}$ as:
    \begin{equation}
        \label{eq:loss_adv}
         L(y,F_{A}(x;\theta')) = -\sum_{i}y_{i}\log F_{A}(x;\theta')_{i}  
    \end{equation}

\subsection{Defense strategy}
The defender's goal is to mitigate the extraction attack threat. Since the attacker relies on the prediction probabilities for a successful attack, in order to defend against it, the defender can carefully modify the \emph{posteriors} (i.e., the prediction probabilities) to maximize the attacker's loss  Eq.~(\ref{eq:loss_adv}).
However, the prediction probabilities $y$ have to keep a certain level of confidence score to maintain the model's utility and transparency.
Hence, we constrain the perturbation magnitude on the posteriors $y$ by budget $\epsilon$ defined as the maximum $\ell_1$ distance between the clean posteriors $y$ and the perturbed ones $y'$, defined as: 
\begin{equation}
    \label{eq:l1-distance}
    \ell_1=||y'-y||_{1} \leq \epsilon
\end{equation}


\subsection{Approach}
\label{sec:approach}



\label{sec:approach}
We propose a new heuristic approach to defend against model stealing attacks that could be applied in both large \ac{CNNs} and Quantized \ac{CNNs} for edge device implementations.
Our defense is characterized by a low computational overhead that does not slow down the model's inference while maintaining a low perturbation of the prediction probabilities. 
In contrast to the state-of-the-art defenses GRAD$^{2}$~\cite{mazeika2022steer} and MAD~\cite{Orekondy2019poisoning}, our technique is not computationally expensive since it does not rely on auxiliary models to simulate attackers. 
Our technique is motivated by the small inference times of both \ac{RS}~\cite{lee2018defending} and \ac{AM}~\cite{kariyappa2020defending} (Table~\ref{tab:compare}). 
We revisit these defenses and we propose a better alternative called \acf{DCP} to overcome their limitations such as high perturbation levels for \ac{RS} or requiring an auxiliary model and deteriorating the model's performance with false positives for \ac{AM}.

Through comparing previous defenses, we describe a general defense mechanism that perturbs the prediction probabilities $F_{V}(x; \theta)$ in a classification problem by:
\begin{equation}
     y' = N( a \cdot F_{V}(x; \theta) + b \cdot r )  \label{eq:defense}
\end{equation}
where $a$, $b$ $\leq 1$ are linear combination parameters to control the proportions of the noise function $r$ and the prediction probabilities $F_{V}(x; \theta)$; $N(.)$ is a sum-to-one normalizer so that $y'$ represents the probability values for the different classes in the prediction problem.

Our strategy in injecting the noise is to decrease the uncertain probabilities. 
Our rationale is that probabilities with low values are of less importance. We assume that if a \textit{well-trained} model is uncertain about the prediction of an input, it is highly probable that the model made a wrong prediction (false positive). 
Hence, modifying these probabilities will not affect the defender's accuracy $Acc(F_{V})$. 
However, if the model predicts an output with a high probability, it is more likely that it is a correct prediction (true positive). Thus, the injected noise is minimal without affecting the defender's performance. 



\begin{figure}[tb]
\centering
\includegraphics[width=\linewidth]{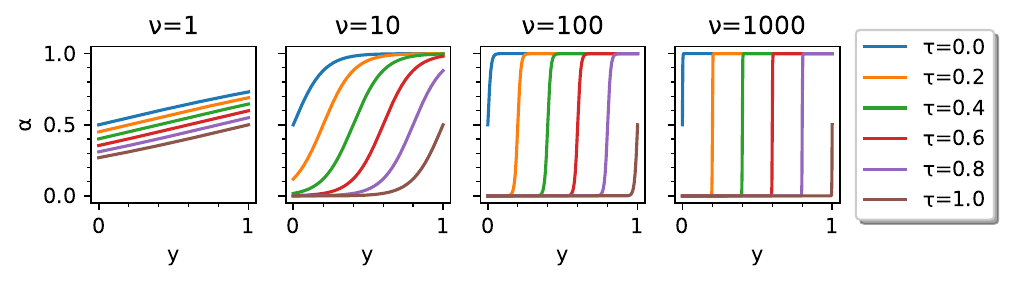}
\caption{The detector function $\alpha$ that switches between the clean and perturbed posteriors using a threshold $\tau$, which pushes $\alpha$ towards $1$ when $y_{max} > \tau$ and towards $0$ otherwise, where $y_{max}$ represents the highest value in a probability vector $y$.
}
\label{fig:switch}
\end{figure}

In order to heuristically find which posteriors to perturb, i.e., low probability values, and switch between clean posteriors $F_{V}(x;\theta)$ and perturbed ones $r(y)$, we leverage a detector function $\alpha$ inspired from \ac{AM}~\cite{kariyappa2020defending} and defined as:
\begin{equation}
\alpha = S (\nu \cdot( y_{max} - \tau ))  
\label{eq:detector}
\end{equation}
where $S$ is the sigmoid function  $S(z)=1/(1+e^{-z})$; 
$\tau$~is a threshold that pushes $\alpha$ towards $1$ when $y_{max} > \tau$ and towards~$0$ otherwise (Fig.~\ref{fig:switch}). Similar to~\cite{kariyappa2020defending}, we set $\nu$ to~$1000$ in order to have a non-smooth switch between clean and perturbed predictions.

In contrast to \ac{AM} that multiplies $F_{V}$ by $(1-\alpha)$ in Eq.~(\ref{eq:dcp}) to create a combination between the clean posteriors and the noise, which results in deteriorating the defender's performance in case of false positives, we set in Eq.~(\ref{eq:defense}) the parameter $a\leftarrow1$ and $b\leftarrow-\beta\cdot\alpha$ 
to obtain: 
\begin{equation}
\label{eq:dcp-small}
     y' = N( F_{V}(x ; \theta) - \beta \cdot   S (\nu \cdot( y_{max} - \tau )) \cdot r(y) ) 
\end{equation}

To compute the perturbation, \ac{AM} leverages another previously trained model, called \textit{misinformation model} $F_{M}(x;\theta)$. The latter is trained on the same data as the victim but in a way that maximizes the loss $ L(y,F_{V}(x;\theta))$. However, this deteriorates the performance of the original model because of false positives.

Instead of using an extra model to inject the noise, our defense leverages a noise function $r(y)$ inspired from \ac{RS}~\cite{lee2018defending}. 
The latter maintains the defender's accuracy in most cases, without being computationally expensive. However, its limitation is that it uses a high perturbation, increasing the $\ell1$ distance between the clean and perturbed predictions.

For simplicity, to avoid estimating the adversary's loss through auxiliary models, 
similar to \ac{RS}~\cite{lee2018defending}, we assume that the attacker has perfectly duplicated the victim model. Thus, we use the victim's predictions $y=F_{V}(x; \theta)$ as starting point to compute the perturbations $r(y)$.
\ac{RS} uses a reverse sigmoid function $S^{-1}(y)=\log(y/(1-y))$ on the predictions $y$ in order to retrieve the initial logits (i.e., the obtained vectors from the last layer). 
Then it selects a perturbation $z=\gamma S^{-1}(y)$, where $\gamma$ is a dataset-specific constant that gives more control over the defense's convergence~\cite{lee2018defending}. 
After that, a sigmoid function is applied on $z$: $S(z)=1/(1+e^{-z})$. Finally, the heuristically obtained perturbation $r$ is:
\begin{equation}
    \label{eq:reverse-sigmoid}
     r(y) = S ( z) - 1/2  = S ( \gamma S^{-1}(y)) - 1/2 
\end{equation}



Our defense's final prediction output is obtained through substituting $r(y)$ in Eq.~(\ref{eq:dcp-small}), 
to finally obtain: 
\begin{equation} 
\label{eq:dcp}
     y'  = N( F_{V}(x ; \theta) - \beta \cdot S (\nu \cdot( y_{max} - \tau )) \cdot (S ( \gamma S^{-1}(y)) - 1/2) ) \\
\end{equation} 

where $S$ and $S^{-1}$ are respectively the sigmoid and reverse sigmoid functions. We set $\tau \leftarrow \beta$ to control the perturbation magnitude.
With this setting we have  $ y' \rightarrow  F_{V}(x; \theta) $ when $ \alpha \rightarrow 0 $  and  $ y'\rightarrow  RS $ when $ \alpha \rightarrow 1$. 

\section{Experiments}
\label{sec:experiments}
In this section, we detail our setup: we explain the datasets, the training process, the quantization techniques, the evaluation metrics as well as the baselines defenses. Then, we report and discuss our results.


\subsection{Setup}
\subsubsection{Datasets}
We tackle \ac{CNNs} trained on 5 benchmark vision datasets with diverse resolutions and different numbers of classes. 
We leverage CIFAR-10~\cite{krizhevsky2009cifar10}, a dataset of images of animals and vehicles with 10 classes; 
CIFAR-100~\cite{krizhevsky2009cifar10}, a diverse real-world images dataset with 100 classes;  
SVHN~\cite{netzer2011svhn} a street view house numbers image dataset with 10 classes; 
and GTSRB~\cite{Stallkamp2011gtsrb}, a German traffic signs dataset with 43 classes. All images are sized to 32$\times$32 pixels.
Furthermore, we work with CUB-200~\cite{WahCUB_200_2011} dataset which contains images of 200 classes of bird species with 224$\times$224 image size.

Since the KnockoffNets~\cite{orekondy2019knockoff} attack requires access to an adversary dataset, we simulate two attack scenarios: a) \textit{knowledgeable adversaries} and b) \textit{knowledge-limited adversaries}. Therefore, to perform the first one, since CIFAR-10 and CIFAR-100 come from the same distribution TinyImages~\cite{krizhevsky2009cifar10}, we select queries $Q$ from CIFAR-10 to attack models trained on CIFAR-100 and vice-versa. Second, for knowledge-limited adversaries we choose unrelated datasets, we select queries from CIFAR-10 as an adversary dataset for victims trained on GTSRB and SVHN. To attack the CUB200 victim, we select images from Pascal-VOC~\cite{Everingham10pascal} which contains a diverse set of objects, such as animals, vehicles, and household items.

\subsubsection{Training}
In order to have a fair comparison with the state-of-the-art techniques and the other baseline defenses, we follow the same training process using the same hyperparameters by the latest technique: GRAD$^2$~\cite{mazeika2022steer}.
For the CUB-200 dataset, we leverage a ResNet-34~\cite{he2016deep} pretrained CNN on ImageNet~\cite{deng2009imagenet} and we fine-tune it for $50$ epochs to obtain our victim model. 
For the other datasets, we train our victim models from scratch for $50$ epochs using the ResNet-18~\cite{he2016deep} architecture.
In KnockoffNets attack~\cite{orekondy2019knockoff}, the adversary architecture is a pretrained WideResNet~\cite{zagoruyko2016wideResNet} for an attack against a CUB-200 dataset victim and ResNet-34 against the small size datasets. In this attack, we limit the query budget to 50k queries.
Similar to~\cite{mazeika2022steer} in training, we use an SGD optimizer with an initial learning rate of $0.01$ that is annealed with a cosine schedule. We use a Nesterov Momentum of $0.9$ and a weight decay of $5\cdot10^{-4}$. 
In DFME~\cite{truong2021data} and MAZE~\cite{kariyappa2021maze} attacks we train adversaries with ResNet-34-8x for small size datasets and WideResNet for CUB-200 dataset. 
We train each model using 20M queries for $200$ epochs. Similar to~\cite{truong2021data}, we use SGD with an initial learning rate of $0.1$ and a weight-decay of $5\cdot10^{-4}$. Additionally, a learning rate scheduler was applied that multiplies the learning rate by a factor of $0.3$ at $0.1\times$, $0.3\times$, and $0.5\times$ the total training epochs. 


\subsubsection{Quantization}
We train \ac{CNNs} for computer vision tasks in our work. Hence, we use two quantization techniques from the literature that are most suited to quantizing our models~\cite{survey_q}: \ac{PTQ} and \ac{QAT}. 
We leverage our previously trained victim models to quantize them with \ac{PTQ}.
We use the same previously described architectures to retrain quantized models with \ac{QAT}.

\subsubsection{Evaluation metrics}
\paragraph{Classification Error and $\ell_1$ distance} 
In order to assess the performance of a defense against stealing attacks, we compute the adversary's classification error (Adv. Clf.Error) on the victim's test set.
This metric measures the stolen functionality from the victim, i.e., the performance of the extracted model on the same task. 
Similar to \cite{orekondy2019knockoff,mazeika2022steer}, we use two utility metrics: the defender's classification error (Def. Clf.Error) and the $\ell_1$ distance between the clean predictions and the perturbed predictions by each defense technique.
The $\ell_1$ distance metric measures the magnitude of the perturbation that impacts the prediction probabilities (Eq.~\ref{eq:l1-distance}). A high $\ell_1$ distance (e.g. $\ell_1 > 1.2$) entails a high perturbation that might harm the model's transparency and reduces its utility when users require confident probabilities.
This helps in identifying the trade-off between the defender's performance and the perturbation budget for each defense.
Both accuracies are calculated using a held-out labeled test set that was not seen before by the victim (during the training) nor the stolen model.

\paragraph{Hardware-related metrics} 
We compute the average inference latency per query and the average GPU energy consumption for all queries using Weights and Biases~\cite{wandb} platform.
This helps perceive the impact of the computation cost of each defense technique in real-world applications.
To have a fair evaluation, we run all experiments on a Linux CentOS~7 computer with 1 GPU NVIDIA Tesla P100-PCIE-12GB and 24 CPU cores Intel E5-2650 v4 Broadwell @~2.2GHz.


\subsubsection{Defenses}
We use several baselines to compare our proposed defenses and validate our approach. \emph{No Defense} means that we attack an undefended model. 
To the best of our knowledge, the best defenses in the literature are: \emph{GRAD$^2$}~\cite{mazeika2022steer}, \emph{MAD}~\cite{Orekondy2019poisoning}, \emph{\ac{AM}}~\cite{kariyappa2020defending}, and \emph{\ac{RS}}~\cite{lee2018defending}.
First, we compare our defense using large (i.e., normally trained) models with all the other defenses.
After that, for the \emph{quantized} models, we compare our defenses only to the techniques that are more suitable for edge device implementations. 
In other words, we select the defense that maintains the inference time without requiring additional models: \emph{\ac{RS}}.




\begin{figure}[]
     \centering
    \includegraphics[width=\linewidth]{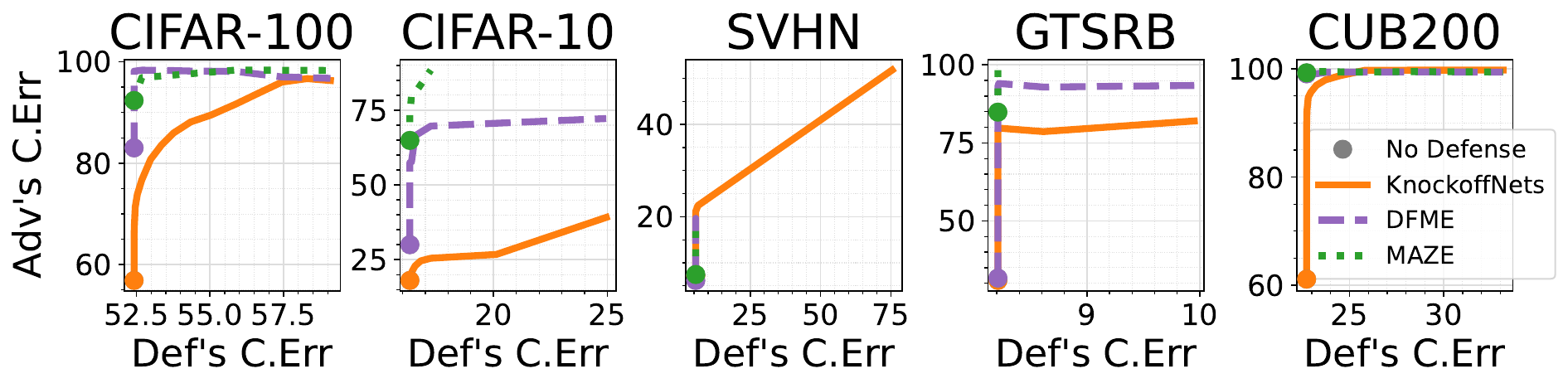}

      \caption{State-of-the-art attacks vs. our defense. Curves are obtained by varying the perturbation magnitude parameter~$\beta$ (Eq.~\ref{eq:dcp}) between $0$ and $1.5$.  
The attack performance is represented on the $y$-axis by the \emph{Adversary's Error}. The $x$-axis considers the \emph{Defender's Error}. 
Our defense is effective in defending against  state-of-the-art stealing attacks.
}
\label{fig:dcp-vs-attacks}
\end{figure}

%

\begin{figure*}[]
     \centering
     \subfloat[Knowledgeable adversary\label{fig:results-b}]{%
  \includegraphics[width=0.36\textwidth]{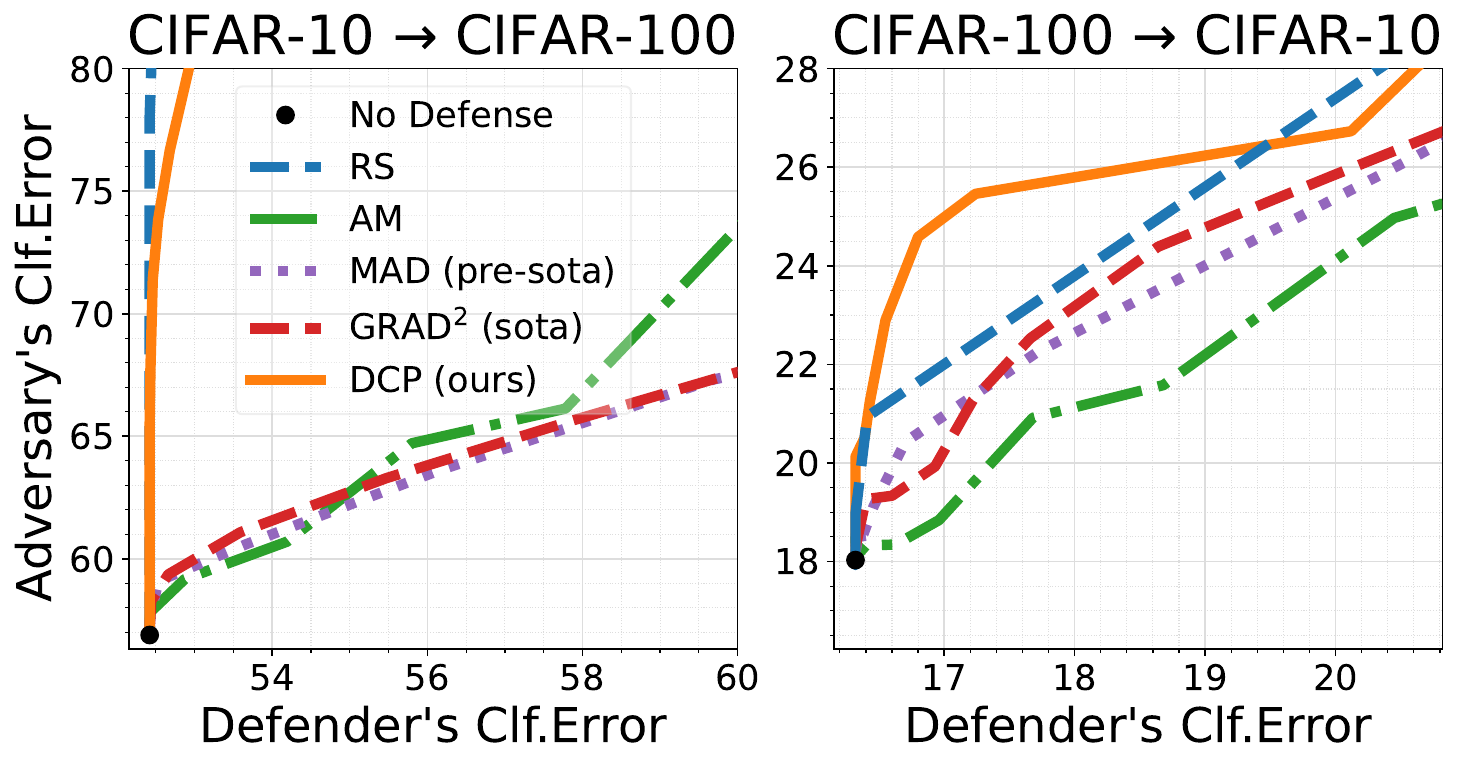}
} \hfil
\subfloat[Knowledge-limited adversary\label{fig:results-a}]{%
  \includegraphics[width=0.54\textwidth]{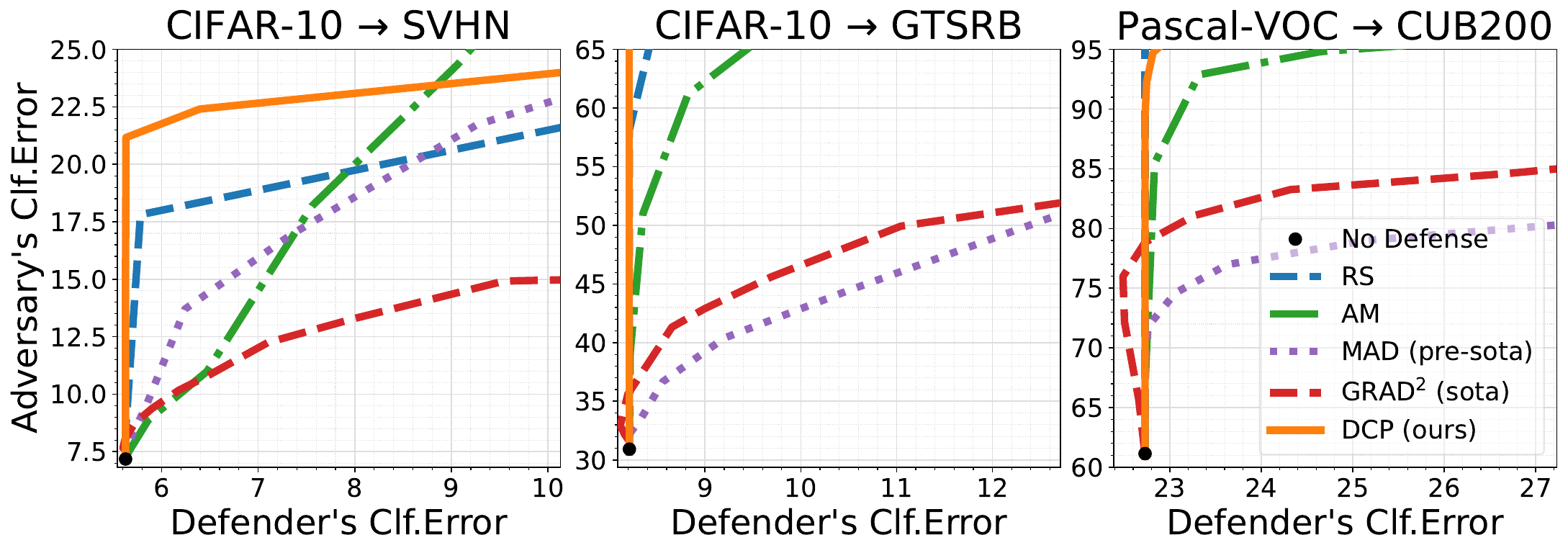}
} \hfil
\caption{
Impact of the KnockoffNets stealing attack \cite{orekondy2019knockoff} on our defense vs. other baselines.
Curves are obtained by varying the perturbation magnitude parameter $\beta$ (Eq.~\ref{eq:dcp}) between $0$ and $1.5$.  
The attack performance is represented on the $y$-axis by the \emph{Adversary's Classification Error}. The $x$-axis considers the \emph{Defender's Classification Error}. 
Our defense outperforms the state-of-the-art defenses on knowledgeable adversaries (a) and knowledge-limited attackers (b).
}
\label{fig:results}
\end{figure*}

\begin{table*}[tb]
\centering
\caption{
Impact of the stealing attack on the defenses with constraints on both the variation of the Defender's Classification Error ``$\Delta$Def.Clf.Error'' and the $\ell_1$ distance between clean and perturbed predictions. 
We report the maximum attainable value under both constraints. 
$\Delta$Def.Clf.Error ranges from $1\%$ to $5\%$ and the $\ell_1$ distance is maintained at a budget of $0.9$. 
\textbf{Bold} denotes the best value and \underline{underlined} denotes the second-best.
}
\begin{tabular}{l l l l l l l l l l l l l l l l l l l l l l l l l} 
 \toprule
\multirow{3}{*}{} & \multicolumn{3}{c}{CIFAR-10$\rightarrow$CIFAR-100}  &  \multicolumn{3}{c}{CIFAR-100$\rightarrow$CIFAR-10} &  \multicolumn{3}{c}{CIFAR-10$\rightarrow$SVHN} &  \multicolumn{3}{c}{CIFAR-10$\rightarrow$GTSRB } &  \multicolumn{3}{c}{Pascal-VOC$\rightarrow$CUB200 }\\  
\cmidrule{2-16}
& \multicolumn{3}{c}{ $\Delta$ Def.Clf.Error} & \multicolumn{3}{c}{ $\Delta$ Def.Clf.Error} & \multicolumn{3}{c}{ $\Delta$ Def.Clf.Error} & \multicolumn{3}{c}{ $\Delta$ Def.Clf.Error} & \multicolumn{3}{c}{ $\Delta$ Def.Clf.Error} \\
Method		&	$1\%$	&	$2\%$	&	$5\%$	&	$1\%$	&	$2\%$	&	$5\%$&	$1\%$	&	$2\%$	&	$5\%$&	$1\%$	&	$2\%$	&	$5\%$&	$1\%$	&	$2\%$	&	$5\%$\\

 \midrule

No defense  	     &	56.89	&	56.89	&	56.89	&	18.03	&	18.03	&	18.03	&	30.92	&	30.92	&	30.92	&	7.18	&	7.18	&	7.18	&	61.14	&	61.14	&	61.14	\\
RS	                 &	\underline{64.20}	&	\underline{64.20}	&	64.20	&	19.22	&	19.22	&	19.22	&	51.07	&	51.07	&	51.07	&	14.23	&	14.23	&	14.23	&	83.80	&	83.80	&	83.80	\\
AM	                 &	59.81	&	61.27	&	\underline{65.87}	&	19.88	&	21.34	&	25.65	&	\underline{58.29}	&	\underline{58.29}	&	\underline{58.29}	&	12.04	&	15.46	&	15.46	&	\textbf{88.92}	&	\textbf{88.92}	&	\textbf{88.92}	\\
MAD (pre-sota)  	 &	60.27	&	61.51	&	64.94	&	21.54	&	23.08	&	\underline{27.24}	&	40.42	&	43.55	&	51.98	&	\underline{14.84}	&	\underline{17.65}	&	\textbf{23.52}	&	77.11	&	78.45	&	80.61	\\
GRAD$^{2}$ (sota) 	 &	60.77	&	62.06	&	65.20	&	\underline{21.56}	&	\underline{23.37}	&	\textbf{27.25}	&	43.73	&	47.27	&	52.51	&	11.15	&	12.86	&	15.02	&	82.04	&	83.49	&	85.36	\\
DCP	(ours)                 &	\textbf{70.31}	&	\textbf{70.31}	&	\textbf{70.31}	&	\textbf{21.87}	&	\textbf{23.78}	&	23.78	&	\textbf{59.73}	&	\textbf{59.73}	&	\textbf{59.73}	&	\textbf{21.69}	&	\textbf{21.69}	&	\underline{21.69}	&	\underline{85.62}	&	\underline{85.62}	&	\underline{85.62}	\\

\bottomrule
\end{tabular}
\label{tab:results}
\end{table*}

\subsection{Results}

We present the results of our experiments in 
Fig.~\ref{fig:dcp-vs-attacks},
Fig.~\ref{fig:results}, 
Fig.~\ref{fig:l1-dcp-vs-rs}, 
Fig.~\ref{fig:queries-budget}, 
Fig.~\ref{fig:results-ptq-qat}, 
Table~\ref{tab:results},
and Table~\ref{tab:inference-energy}. In the figures, we visualize two utility metrics: the defender's error and the $\ell_1$~distance between the original predictions $y$ and the perturbed ones $y'$. The attack performance is measured by the adversary's classification error. The higher this error, the stronger the defense.

\subsubsection{\ac{DCP} defense vs. state-of-the-art attacks} 
Fig.~\ref{fig:dcp-vs-attacks} presents an evaluation of our defense \ac{DCP} (Eq.~\ref{eq:dcp}) against the three attacks KnockoffNets~\cite{orekondy2019knockoff}, DFME~\cite{truong2021data}, and MAZE~\cite{kariyappa2021maze}. We observe the performance of these attacks on undefended and defended models. 
A good adversary produces a model with a low classification error. 
As shown in Fig.~\ref{fig:dcp-vs-attacks}, using our defense approach for all datasets and attack models can significantly reduce the effectiveness of the attacker by increasing its error.
Our defense produces reasonable operating points across all models, with a minimal impact on the defender's error and a high impact on the adversary's error. 
For instance, a defended victim trained on the SVHN dataset can multiply the adversary's error by up to $\times3.3$.
From Fig.~\ref{fig:dcp-vs-attacks} we notice that the strongest model stealing attack is KnockoffNets as it produces the lowest adversary's error for most datasets. 

\subsubsection{\ac{DCP} defense vs. baselines} 
We evaluate our defense and the baselines against the strongest stealing attack in the attack models evaluation: KnockoffNets.
Fig.~\ref{fig:results} shows that for knowledgeable and knowledge-limited attackers, our defense technique outperforms the state-of-the-art defenses in most cases. 
It increases the adversary's classification error significantly before it starts impacting the defender's classification error.
For instance, for CIFAR-10 (Fig.~\ref{fig:results-b}, right column), our defense \ac{DCP} outperforms other techniques in increasing the adversary's classification error by $22\%$ with a small impact on the defender's performance ($1\%$).
Most baseline defenses fail to decrease the adversary's performance without deteriorating the defender's performance. 
For a victim trained on GTSRB, when we constrain only the defender's error to be lower than $1\%$, our defense \ac{DCP} increases the adversary's classification error by more than $109\%$.
The second-best is \ac{RS} which increases the adversary's classification error up to $79\%$.

In order to balance both metrics measuring the impact on the defender's performance and the $\ell_1$ distance between clean and perturbed predictions, we constrain both the $\ell_{1}$ distance with a maximum budget of $0.9$ and the defender's error variation to be lower than $1\%$, $2\%$ or $5\%$. 
Table~\ref{tab:results} shows the results of this experiment. In most cases, our defense \ac{DCP} yields comparable to or better results than the state-of-the-art defenses while offering reasonable trade-offs that prioritize the defender's performance.

Through these extensive experiments, we prove that our results are significantly better than the state-of-the-art for knowledgeable and knowledge-limited adversaries. 
We provide better robustness against stealing attacks while maintaining a reasonable perturbation.


\subsubsection{Ablation study} 
\begin{figure}[]
     \centering
  \includegraphics[width=0.9\linewidth]{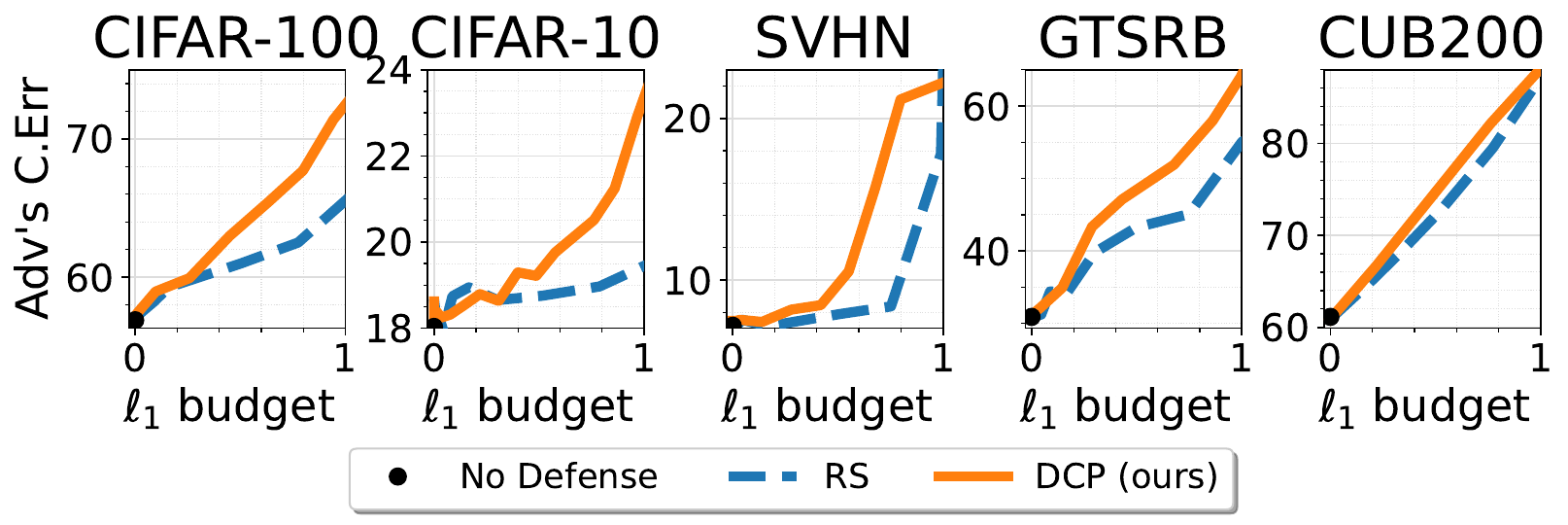}%
\caption{
Our defense vs. Reverse Sigmoid. The Adversary's classification error with relation to the $\ell_1$ distance between the perturbed predictions and the clean predictions. 
Our defense achieves higher performance using lower perturbation budget.
}
\label{fig:l1-dcp-vs-rs}
\end{figure}

\paragraph{Comparing $\ell_1$ distance of DCP and RS}
In the previous experiments, we observe that RS~\cite{lee2018defending} and our defense DCP sometimes achieve the same attack performance, i.e., they both increase the adversary's error to comparable values with a fast inference latency. This could be explained by the fact that our defense uses a noise function that draws some inspiration from RS. We evaluate both defenses from a perturbation magnitude standpoint. Fig.~\ref{fig:l1-dcp-vs-rs} shows that our defense outperforms RS in the matter of $\ell_1$ budget. In fact, our defense uses less perturbation to increase the adversary's classification error. Hence, DCP overcomes the limitation of RS that requires a high perturbation to perform well.
 
\begin{figure}[]
\centering
\includegraphics[width=1\linewidth]{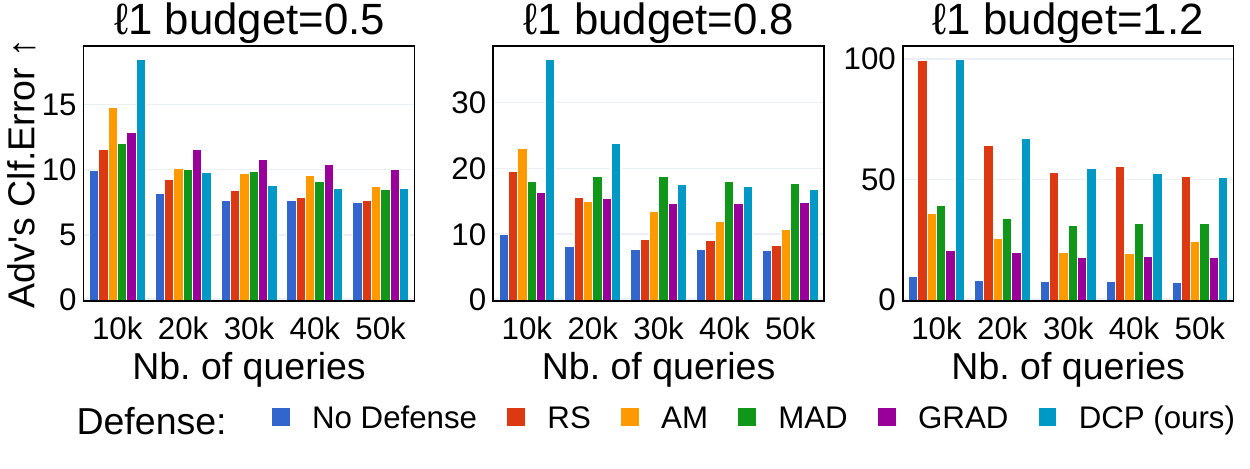}
\caption{Performance of the defenses against extraction attacks relying on different numbers of queries. Each column considers $\ell_{1}$ budget constraint. The attacks are performed on a victim trained on the SVHN dataset. In most cases of a low number of queries, our defense outperforms other techniques in increasing the adversary's classification error.
}
\label{fig:queries-budget}
\end{figure}

\paragraph{The impact of the number of queries used in the stealing attack}
We explore the resilience of our defense in protecting models from multiple stealing attacks with different budgets of queries.
For each attack, we select a budget for the number of queries ranging from $10k$ to $50k$ images.
Each victim model is protected with a perturbation-based defense constrained by an $\ell1$ budget for the distance between the clean and perturbed predictions.
Fig.~\ref{fig:queries-budget} shows the results of this experiment on the SVHN dataset.
Our defense increases the adversary's classification error significantly higher than other defenses, especially for a lower number of queries. 
This entails that our defense slows down the adversary which makes him use a larger number of queries to perform the attack.

\subsubsection{Inference latency and energy consumption}

To validate the efficiency of our proposed defense, we report in Table~\ref{tab:inference-energy} the hardware-related evaluation metrics for our defense and the baselines.
For example, for the CIFAR-10 dataset, we obtain  $\times37$ faster inference than 
the state-of-the-art technique
GRAD$^2$
 and more than $\times527$ faster inference than 
 MAD.
In addition, we show that our defenses have comparable minimal energy consumption for each dataset compared to other defenses.  
Finally,  we only require half the memory used by most of the baselines, since our technique does not require an extra model to compute the perturbations.
To conclude, our defense is practical for edge device implementations: in general, we report comparable-to-best results in inference latency and energy consumption.


\begin{table}[]
\centering
\caption{The average inference latency per query and the average energy consumption required to 
generate all perturbed queries, \textbf{bold} means best.
}
\label{tab:inference-energy}
\begin{tabular}{cccccc}
\toprule
 & CIFAR10 &	CIFAR100 &	SVHN &	GTSRB &	CUB200\\
\cmidrule{2-6}
Defense & \multicolumn{5}{c}{Inference latency ($ms$)} \\
\midrule
No defense    &	0.10  &	0.09  &	0.09  &	0.07    &	0.09            \\

RS \cite{lee2018defending}   &	\textbf{0.10} & 0.10  &	\textbf{0.09} &\textbf{0.13}  &	0.10  \\

AM \cite{kariyappa2020defending} &	0.14   &	0.14  &	0.13 	& 0.16 &	0.17  	\\

MAD  \cite{Orekondy2019poisoning}	& 58.02  &	499.09	 & 59.12 	& 225.50   &	121.4  \\

GRAD$^2$  \cite{mazeika2022steer} &	4.09 	& 21.83  &	4.39 	& 14.47   &	44.17   	\\


DCP (ours)	& 0.11 	& \textbf{0.09}  &\textbf{0.09}  & 	\textbf{0.13}  	 &	\textbf{0.09}   \\


\midrule
Defense  & \multicolumn{5}{c}{ Energy consumption ($Wh$)} \\
\midrule
No defense     &	 0.13 &	0.15 &	 0.20 &	 0.12    &	 0.43            \\
RS \cite{lee2018defending}     &	 0.18	&  \textbf{0.15} &	 \textbf{0.20}&\textbf{0.12} & 0.90  \\
AM \cite{kariyappa2020defending}    &	 0.21 &	0.18 &	 0.23	&  0.17 &	 0.93  	\\
MAD  \cite{Orekondy2019poisoning}  	& 2.79 &	 22.18 &  6.96	&  12.66  &	 80.06  \\
GRAD$^2$  \cite{mazeika2022steer}     &	 0.42	& 1.20 &	0.94	&  1.19  &	 12.15  	\\
DCP (ours)	&  \textbf{0.17}	& 0.17 & \textbf{0.20} & 	 \textbf{0.12} 	 &	 \textbf{0.59}  \\

\bottomrule
\end{tabular}
\end{table}

\begin{figure}[]
\centering
\subfloat[PTQ\label{fig:ptq}]{%
  \includegraphics[width=0.38\textwidth]{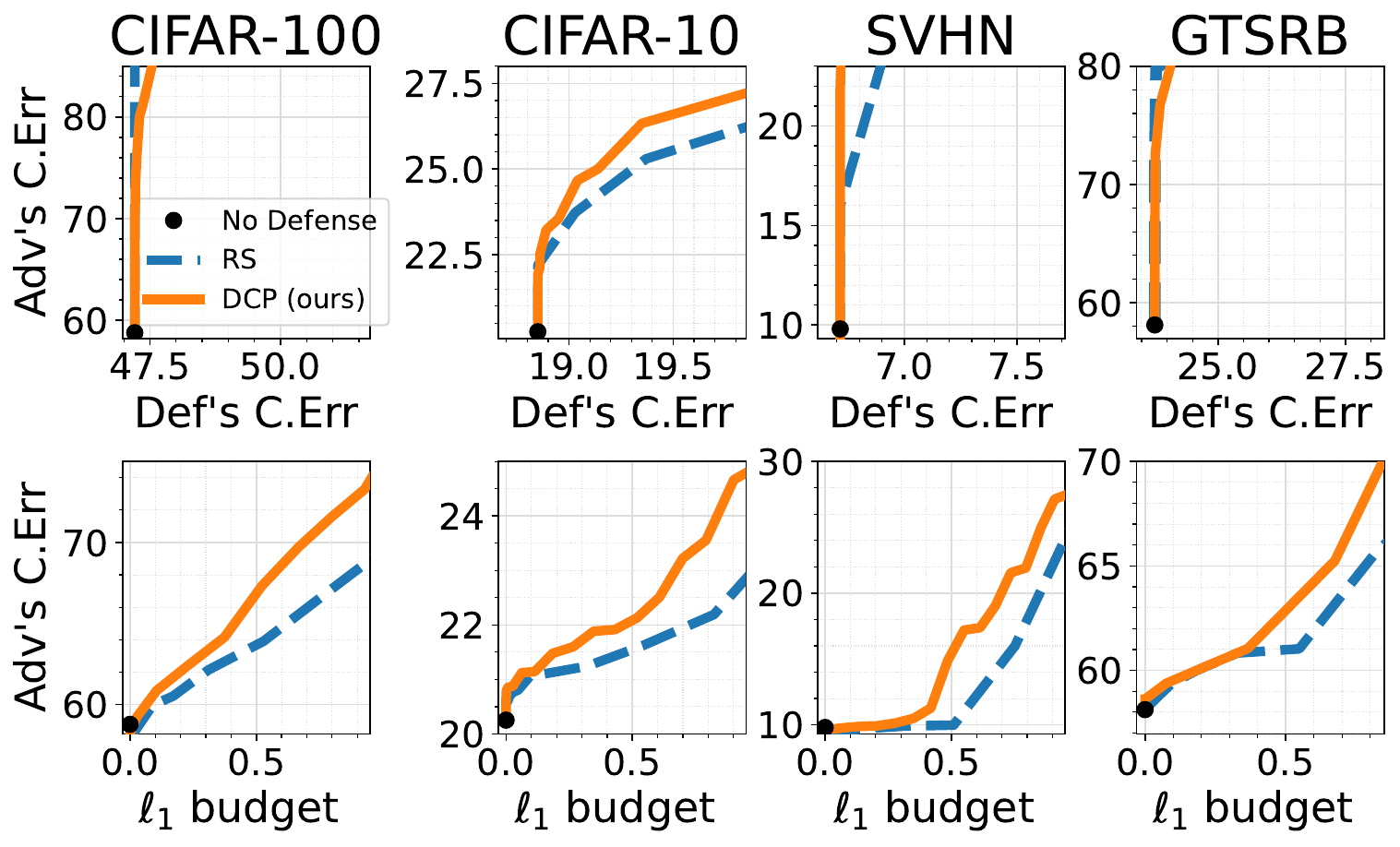}%
} \hfil
\subfloat[QAT\label{fig:qat}]{%
  \includegraphics[width=0.38\textwidth]{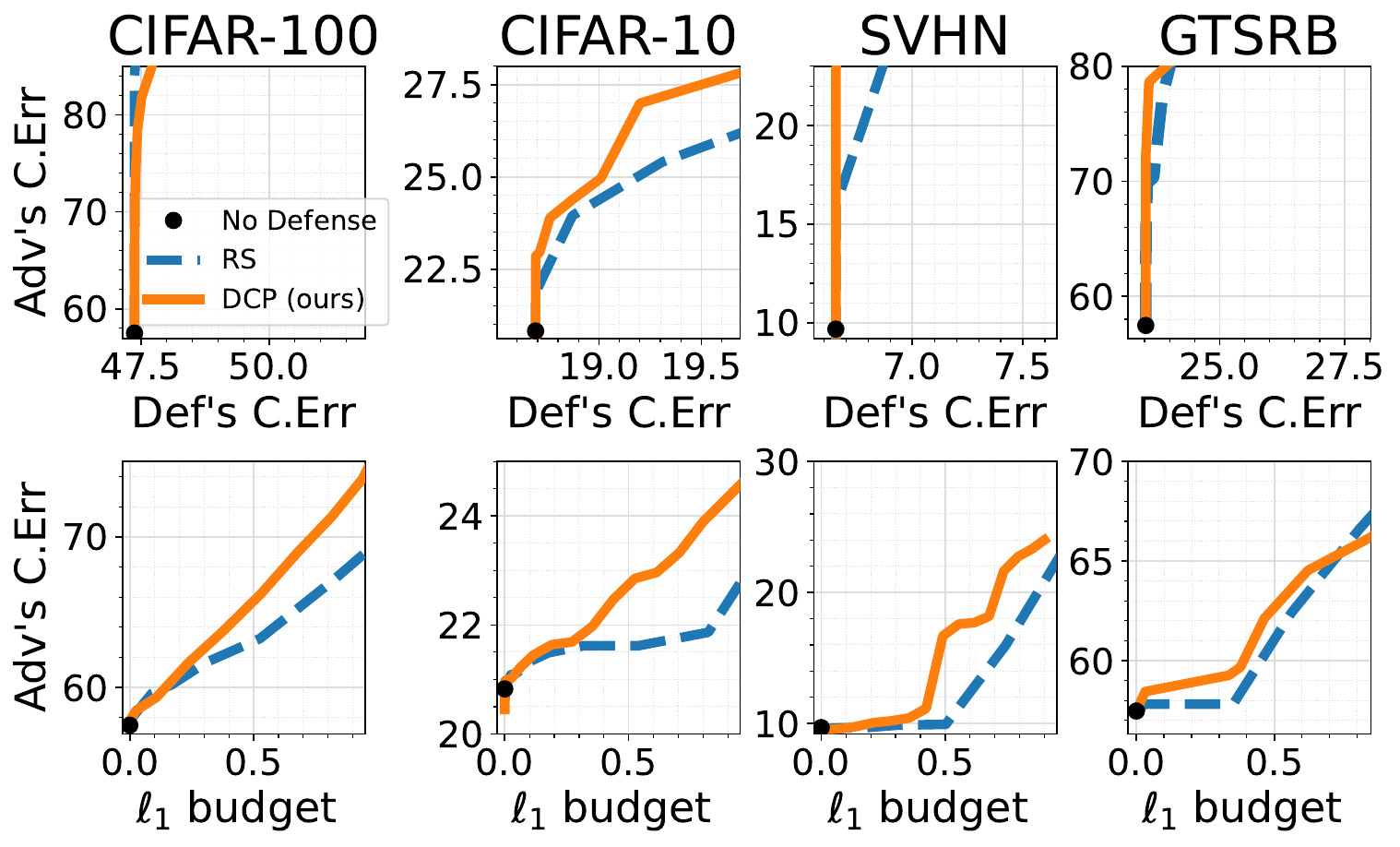}%
} \hfil
\caption{Performance of the defense against stealing attack on Quantized \ac{CNNs} with two quantization techniques PTQ (a) and QAT (b). Our defense outperforms the baseline defense that does not require an auxiliary model, nor increases inference delays. 
}
\label{fig:results-ptq-qat}
\end{figure}

\subsubsection{Model stealing defenses on quantized models}

We validate the extension of our techniques on quantized \ac{CNNs}. Fig.~\ref{fig:results-ptq-qat} shows that in most cases our defense outperforms or matches the results of RS. For example, in PTQ results (Fig.~\ref{fig:ptq}), for a selected perturbation budget (e.g. $\ell_1=0.5$), our technique \ac{DCP} provides better protection against model stealing attacks through increasing the adversary's classification error to between $5\%$ and $20\%$ while maintaining the defender's performance.
Our technique sometimes achieves the same performance while having lower perturbations. This shows that our defense can be used in \ac{DL} models targeting embedded systems applications.

\section{Conclusion}
\label{sec:conclusion}
The motivation for our work is the non-efficiency of previous defense techniques against model stealing attacks. This hinders their deployment in edge devices. We overcame the heavy computation problem with a novel heuristic defense.
We empirically proved that it works efficiently on large \ac{CNNs} and their corresponding quantized versions targeting edge devices. 
We found that our approach achieves comparable-to-better performance with the state-of-the-art while maintaining low perturbations and fast inference time. 
Our defense proved to be effective in amplifying the adversary's classification error under various attack models.

With this paper, we shed light on the need for considering the hardware limitations during the development of security and privacy algorithms for \ac{DL} applications.
We strongly encourage the community to prioritize the 
practicality of their proposed defenses, not as an undesired side-effect, but as a crucial requirement for real-world applications.
For future work, we intend to work
on developing efficient defenses specifically tailored for edge devices and real-time systems.

\bibliography{document}
\bibliographystyle{IEEEtran}			

\end{document}